\documentclass[conference]{IEEEtran}
\IEEEoverridecommandlockouts
\usepackage{cite}
\usepackage{amsmath,amssymb,amsfonts}
\usepackage{algorithmic}
\usepackage{graphicx}
\usepackage{textcomp}
 \usepackage{subcaption}
 \usepackage{url}
 \usepackage{xurl}
 \usepackage{hyperref} 

\usepackage{xcolor}
\def\BibTeX{{\rm B\kern-.05em{\sc i\kern-.025em b}\kern-.08em
    T\kern-.1667em\lower.7ex\hbox{E}\kern-.125emX}}
    \makeatletter
\newcommand{\linebreakand}{%
  \end{@IEEEauthorhalign}
  \hfill\mbox{}\par
  \mbox{}\hfill\begin{@IEEEauthorhalign}
}
\makeatother
\begin{document}

\title{An offline approach to fNIRS-guided reinforcement learning for robot behavior}

\author{\IEEEauthorblockN{Julia Santaniello$^1$, Madelaine Brower$^1$, Benson Jiang$^1$, Donatello Sassaroli$^1$, Chenyuan Zhang$^2$, 
\\Robert Jacob$^1$, Jivko Sinapov$^1$}
    \IEEEauthorblockA{\textit{$^1$ Tufts University, School of Engineering}}
    \textit{$^2$ New York University, Tandon School of Engineering}}

\maketitle

\begin{abstract}
Human-in-the-loop Reinforcement Learning has become a popular approach for training, finetuning, and aligning robot behavior with user preferences. Our paper explores the feasibility of using brain signals via functional near-infrared spectroscopy (fNIRS) to modulate robot learning in simulation. We compare agents trained on passive (observational) versus active (demonstrative) interaction tasks, and test multiple methods for enhancing the RL algorithm with the neural signal, focusing on parameter augmentation in contrast to replacement. We further examine how model granularity and noise affect agent learning. Our results show that this framework is effective. The neural signal improves learning when augmenting trajectory priorities and state-action q-targets. Additionally, the framework learns successfully from offline data, offering a practical alternative for settings where real-time BCI setups are impractical or only limited data is available.
\end{abstract}
\section{Introduction}
Human-in-the-loop Reinforcement Learning (HITL-RL) is a growing approach for training and finetuning autonomous agents \cite{retzlaff}. This practice sits within reinforcement learning (RL), a branch of machine learning in which agents learn to take actions by maximizing a reward function, dependent on the state of their environment \cite{sutton}. Existing HITL-RL techniques include learning from demonstration \cite{schaal}\cite{hester}, preference learning \cite{preference}, and intervention learning \cite{intervention}. 

Many of these approaches share the common trait of requiring explicit feedback from human users, generally through speech, physical demonstration, or preference expression. This approach can introduce a few issues. For example, sustained attention and the need to provide expert feedback may lead to cognitive overload for the human teacher \cite{bias}. Second, these systems may introduce bias by assuming that a user's implicit reactions follow the general distribution of the training set. They assume users move, speak, or interact with an agent in a similar manner to others \cite{casper}. These methods can exclude people with motor or speech impairments from finetuning agent behavior.

Implicit HITL-RL loosens the first problem by using passive feedback from users to convey knowledge to the agent. Unlike explicit feedback, where users attempt to shape the policy through direct instruction, implicit systems utilize indirect human reactions to guide agent behavior through gesture, facial expression, or voice inflection \cite{empathic} \cite{dqn-tamer}. However, these methods do not address the second problem, as they generally require users to be able to move and speak. 
\begin{figure}[t]
\centerline{\includegraphics[width=\columnwidth]{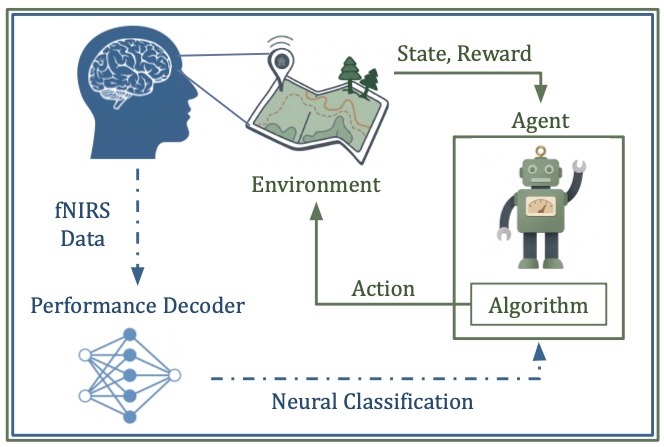}}
\caption{\textbf{NEURO-LOOP:} A user observes an agent taking actions in its environment. Neural data is recorded, preprocessed and then decoded into a value that reflects agent performance. This signal updates the agent's learning algorithm.}
\label{fig}
\end{figure}

Our work directly targets this issue by using brain signals as the main feedback channel directing the agent. Sometimes referred to as \textit{intrinsic feedback}, this feedback signal is implied from human biosignals. Emerging work in this domain has introduced encephalography-based (EEG) feedback as a binary signal for training agents via RL \cite{poole}. Instead, we use functional near-infrared spectroscopy (fNIRS), which captures a slower hemodynamic signal with greater spatial resolution, and can be more robust to certain artifacts that plague EEG \cite{pinti}. We propose using fNIRS and EEG in tandem, as opposed to in replacement, as both signals offer complementary features that can be useful for future multimodal, hybrid systems.

We use fNIRS data from a prerecorded dataset, and use an offline approach (not real-time) to guide a simulated robot's policy via RL. We map fNIRS feature vectors to a label describing agent optimality, and use this neural classification to guide the robot's training loop.

Our key contributions are:
\begin{itemize}
    \item We propose, develop, and test a framework for integrating offline neural data into an RL loop, examining multiple methods of algorithm enhancement.
    \item We demonstrate this framework using offline fNIRS data and show improvement in success rate and total episode returns, suggesting that the fNIRS signal is useful without relying on a real-time system, making this approach more deployable in settings where an online pipeline is not practical.
    \item We conduct an ablation study that teases apart the contributions of individual and combined augmentation techniques, suggesting that  Q-value and Priority augmentation perform best, yet augmenting Reward has no effect.
\end{itemize}

We further evaluate how classifier noise effects learning outcomes with a small ablation study. This study allows us to characterize how much noise the framework can tolerate before performance considerably worsens.
\section{Related Work}
HITL-RL is a growing approach using human feedback to guide agent training \cite{retzlaff}. Our approach leverages passive fNIRS signals as feedback, letting users interact with an agent without speaking or moving.
\\
\textit{Human-in-the-Loop RL}: 
Early HITL-RL approaches used binary evaluative feedback to guide learning. \cite{tamer} introduced a framework letting non-expert users guide agent training via evaluative feedback (e.g., good, bad). \cite{tamer-rl} extends these ideas to deep RL, integrating feedback into various parts of the algorithm to shape policy. We adopt some of these ideas, augmenting the reward, Q-value, and priority of an MDP tuple with the neural signal at a given timestep.
\\
\textit{Implicit Human-in-the-Loop RL}:
Implicit HITL-RL shares this framework, but feedback takes the form of subliminal expressions not directed at the agent. \cite{empathic} used facial expressions and gestures to update an agent's policy in real-time, decoding expressions into learning task statistics before applying them to the RL loop. \cite{dqn-tamer} similarly uses decoded emotional states ("happy," "sad") as positive or negative reward.
\\
\textit{EEG for Intrinsic HITL-RL}:
Recent work has explored implicit brain signals, particularly EEG-measured error-related potentials (ErrPs), as HITL-RL feedback \cite{eeg-rl-luo, eeg-rl-wang, agarwal2, kim}. ErrPs are binary signals detected via EEG when a person perceives an error, evaluated in game domains like Maze and MuJoCo's Ant and Cheetah \cite{agarwal2, eeg-rl-luo}, and in simulated robot domains \cite{kim, eeg-rl-aki, eeg-rl-wang}. Most approaches use this binary signal to augment or replace a sparse reward function. We extend this approach through using the fNIRS hemodynamic response as our neural signal, and augmenting Q-value and trajectory prioritization on top of reward, learning from a limited, offline dataset.
\section{Theoretical Framework}
\subsection{The Reinforcement Learning Problem}
A reinforcement learning agent learns a task in OpenAI Gymnasium's Fetch Pick-and-Place domain (v4), a robotic manipulation environment structured as a Markov Decision Process (MDP), described by the tuple $(\mathcal{S}, \mathcal{A}, \mathcal{P}, \mathcal{R}, \gamma)$. $\mathcal{S}$ is the set of observable states, $\mathcal{A}$ the set of actions, and $\mathcal{P}$ is the transition function that determines the next state given the state and action at time $t$: $P(s_{t+1}|s_t,a_t)$. $R$ denotes the reward function, and $\gamma$, the discount factor, modulates the weight of future rewards.

\subsection{The Performance Decoder}
Previous work has mapped human brain signals to various granularities of agent performance \cite{me-rldm, me}. We adopt this approach to test whether neural data decoded into a value that reflects action optimality can improve a robot's policy by modulating the RL algorithm. We denote the Multilayer Perceptron (MLP) performance decoder as $\mathcal{M}_g(f_t) = \hat{y}_t$, where $f_t$ is the neural feature vector at time $t$, $g$ the granularity of the optimality label, and $\hat{y}_t$ the predicted class.
We denote the confidence of the binary or ternary classifier as $c_{t} = P(y_t = \hat{y}_t \mid f_t)$: the probability that predicted class $\hat{y}_t$ matches true class $y_t$, given $f_t$. For regressors mapping neural features to a continuous label representing action-selection error, we denote the parallel quantity as $c_{t} = 1-\hat{e}_t$, the estimated probability that the agent chose a near-optimal action given $f_t$.
\section{Methods}
 \begin{figure*}[t]
\centerline{\includegraphics[width=\textwidth]{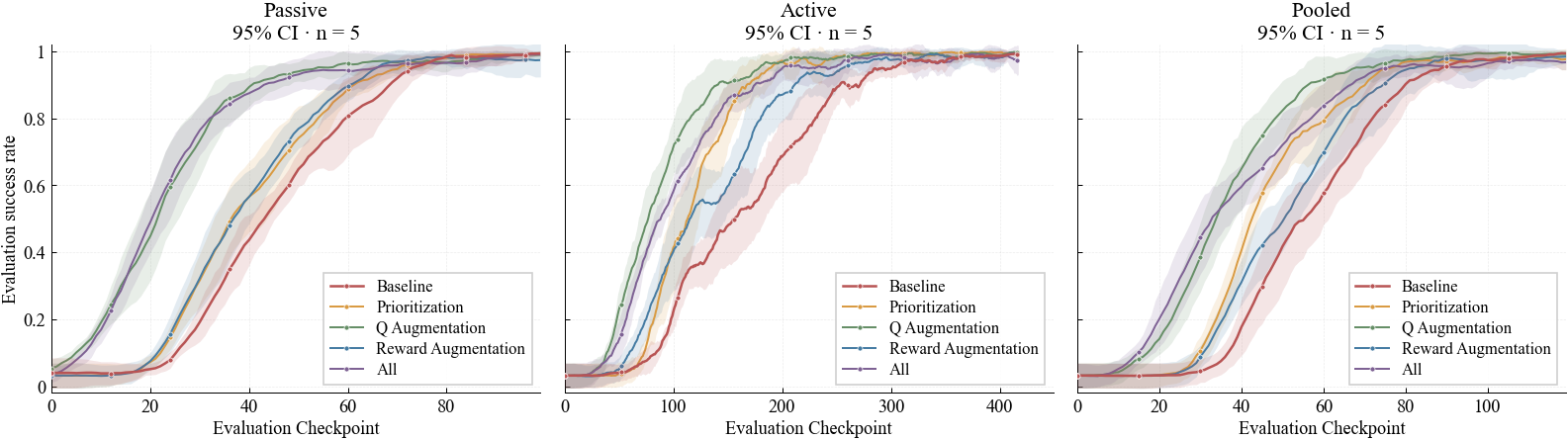}}
\caption{\textbf{Robot Learning Curves (Binary):} Success rate curves of each augmentation method across interaction tasks. \textit{Q-augmentation} and \textit{All} conditions perform best. \textit{Prioritization} outperforms baseline, but modestly. \textit{Reward Augmentation} performs similarly to \textit{Baseline}. 95\% confidence intervals are plotted as error bars. Evaluation checkpoints occur every 10,000 training steps, and each curve is averaged across 5 trials, $n$. Beta $=$ best; Finetune Checkpoint $= 0.0$.}
\label{fig:best_settings}
\end{figure*}
Our framework, NEURO-LOOP, is comprised of two major components: (1) the neural classification problem and (2) the RL loop. \cite{me} addresses the neural classification problem, presenting three performance decoder variations that map passive fNIRS feature vectors to agent performance labels. This section covers the second component which uses the performance decoder's output to guide the RL algorithm.
\\
\textbf{Goal:} The robot's goal is to find, pick up, and place a black cube in a randomized location on the table in front of it.
\\
\textbf{DDPG-HER with Prioritization:} We train the robot with Deep Deterministic Policy Gradient with Hindsight Experience Replay (DDPG-HER) \cite{her}. We further implement \cite{schaul}-like prioritization, where batched experiences are given priority instead of being uniformly sampled. We chose this algorithm because of its documented success with this specific task.
\\
\textbf{Finetuning Checkpoints:} We inject the prerecorded MDP tuples at success rate checkpoints determined at evaluation time. The agent learns from scratch, and once it passes this checkpoint, it learns from the dataset until completely used, then resumes online learning until the total episode count is reached. Our goal is to determine whether this framework is better suited for early, mid or late-stage training.
\\
\textbf{Interaction Tasks:} We record and utilize passive fNIRS data from a participant's prefrontal cortex during one of two interaction tasks.
\\
\textit{Passive:} The participant only \textbf{observes} the agent. State-action pairs are chosen by a near-optimal policy that is forced to execute a mix of optimal and suboptimal actions (roughly 50\% each), so every MDP tuple carries a known performance label.
\\
\textit{Active:} The participant \textbf{teleoperates} the robot, so the state-action pairs aligned with fNIRS data are those the participant selected via controller. Since participants were not expert demonstrators, active MDP tuples are commonly suboptimal. \textit{Pooled} experiments combine active and passive data.
\\
\textbf{Decoder Granularity:}
Training a performance decoder requires choosing the granularity of the performance label, $g \in \{b, d, e\}$. \textit{Binary decoders} predict Optimal or Not Optimal, $g_b \in \{0, 1\}$. \textit{Ternary decoders} predict Optimal, Suboptimal, or Not Optimal, $g_d \in \{0, 1, 2\}$. \textit{Continuous decoders} are regressors that predict the error between the agent's chosen action and actions chosen by near-optimal policies, $g_e \in \mathbb{R}$.
\\
\textbf{Tuning the Performance Decoder Output:}
We train a performance decoder for every combination of interaction task and decoder granularity, using all available participant data in a multi-subject training paradigm. We then tune the neural classification to align with the signal it augments.  The regressor output is recalibrated as: $n_t = (x_t + c_{t} \cdot \mu_X[0])\cdot \beta$ and the classifier output: $n_t = (x_t + c_{t} \cdot \mu_X[\hat{y_t}])\cdot \beta$, where $\hat{y}$ is the raw decoder output, $x_t$ the parameter to be augmented (Reward, Priority, Q-value), and $\mu_X$ is an array that describes the expected value of some parameter for each optimality class. From here, $n_t$ is what is applied to the RL loop.
\\
\textbf{Augmentation Techniques:} Conditions build on DDPG-HER with Prioritization, and differ only in how $n_t$ modulates the RL loop. We chose three different methods that operate well with offline learning.
\\
\textit{Baseline:} Data is injected at the finetuning checkpoint, but the neural augmentation step is withheld.
\\
\textit{Prioritization:} $n_t$ modulates trajectory priority. TD-error $\delta$ is the fallback priority for all non-Prioritization conditions.
\\
\textit{Reward Augmentation:} $n_t$ modulates the domain's reward.
\\
\textit{Q-Augmentation:} $n_t$ modulates the state-action Q-value.
\\
\textit{All:} Combines Prioritization, Reward, and Q-Augmentation.
\\
\textbf{Decoder Noise:} We inject synthetic noise into the performance decoder's output to test its effect on learning. With probability $\tau$, classifier outputs are flipped, or regressor outputs are shifted by a random value between 0 and 1.
\section{Experiments and Results}
\noindent
\textbf{Dataset:} We evaluate \textit{NEURO-LOOP} on a publicly available dataset that pairs fNIRS recordings with the behavior of an autonomous agent. We focus on the Robot domain, where participants either observe or teleoperate a robot that is performing a fetch-and-place task. The dataset provides an MDP tuple consisting of the agent's state, action, and reward, along with the participant's brain signal at each timestep. Each MDP tuple includes three labels describing how optimal the robot's selected action was.
\begin{figure}[t]
\centering
\begin{subfigure}{\columnwidth}
    \centering
    \includegraphics[width=\columnwidth]{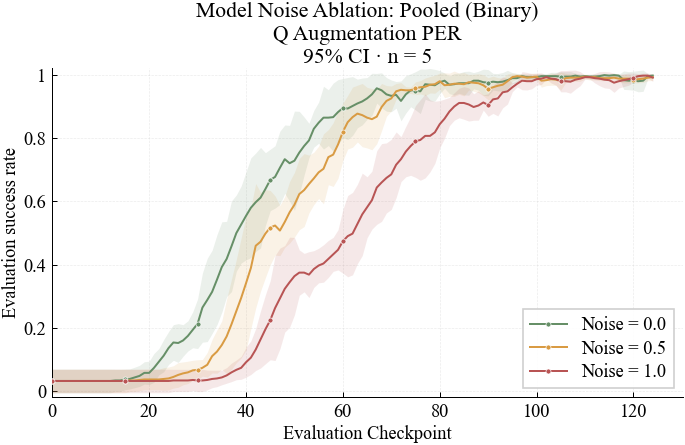}
    \label{fig:sub1}
\end{subfigure}
\vspace{0.5em}
\begin{subfigure}{\columnwidth}
    \centering
    \includegraphics[width=\columnwidth]{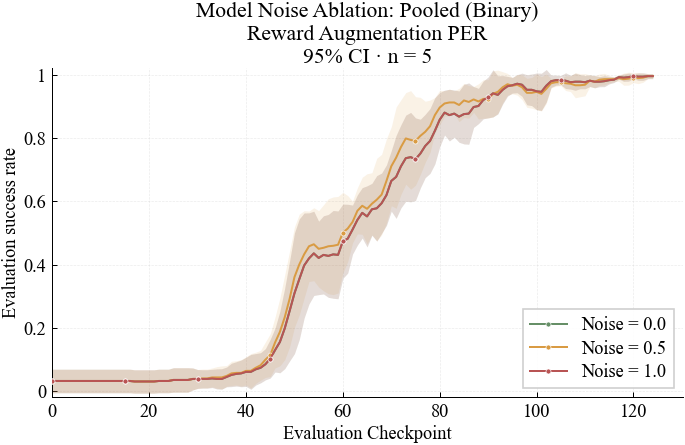}
    \label{fig:sub2}
\end{subfigure}
\caption{\textbf{Decoder Noise:}  We inject synthetic noise into the performance decoder's output to test its effect on learning. Conditions that perform considerably above \textit{Baseline} (Q-Aug) degrade as noise increases, while conditions that perform similarly to \textit{Baseline} (Reward Aug) are unaffected. This pattern suggests that policy improvements are a direct result of the neural signal itself: as noise increases, gains decrease back towards \textit{Baseline}. Beta $=1.0$; Finetune Checkpoint $= 0.0$.}
\vspace{-7pt}
\label{fig:noise}
\end{figure}
\\
\textbf{Participants:} We analyzed brain data from 21 participants (11 female, 10 male), ages 19--27. Each participant completed one or both of two possible interaction tasks with at least 13 participants per condition.
\\
\textbf{Metrics:} We tracked the agent's total episode return and success rate at each evaluation cycle. Evaluations were run every 10{,}000 training steps at 25 episodes per evaluation.

\subsection{Augmentation Methods}
\noindent
\textit{Q-Augmentation} and \textit{All} consistently outperformed  \textit{Baseline} and all other conditions (Fig. \ref{fig:best_settings}). We believe this is because \textit{Q-Augmentation} most directly influences the policy. 
\\
\textit{Prioritization} ranked third, showing a modest but consistent improvement over \textit{Baseline} across most interaction tasks. This may be because \textit{Baseline} is already prioritizing trajectories with TD-error, so \textit{Prioritization} is competing with an already well-established ranking scheme. 
\\
\textit{Reward Augmentation} showed little to no improvement, performing similarly to \textit{Baseline}. These results are consistent with that of \cite{kim} finding that applying neural feedback to the agent's reward pipeline performed similarly to agents learning from a dense reward function. All reported differences use 95\% confidence intervals, shown as error bars.

\subsection{Decoder Granularity}
\noindent
\textit{Passive Experiments: }Continuous outperformed binary and ternary in \textit{Q-Augmentation}. Binary and continuous performed similarly elsewhere. Ternary underperformed overall but still outpaced \textit{Baseline} in most conditions.
\\
\textit{Active Experiments:} Binary and ternary outperformed continuous, which only marginally outpaced \textit{Baseline} in \textit{All}.
\\
\textit{Pooled Experiments:} Binary performed best while continuous and ternary struggled. However, ternary \textit{All} and continuous \textit{Prioritization} and \textit{Q-Augmentation} still modestly outpaced \textit{Baseline}.
We theorize ternary underperformed due to noisier output from the performance decoder: F1 scores were ~0.47--0.50 for three-class decoders versus 0.67--0.72 for binary. Consistent with this train of thought, setting decoder noise to 0.5 made Passive and Pooled binary performance erode to that of ternary.

Continuous granularity performed worst in Active, possibly because demonstration data accumulates error, making an error-based signal harder to parse and learn from. Ternary, despite generally underperforming, performed similarly to binary in Active tasks. Across granularities, Prioritization surprisingly showed little variance in performance. This could possibly be because, as a sort of ranking system, separating trajectories into "good" or "bad" is sufficient, with magnitude adding little more. It would be interesting to test whether this pattern changes with alterations to the algorithm's memory buffer.

\subsection{Interaction Task}
As expected, Passive interaction tasks led to the fastest learning. This is likely because Passive tasks maintain a roughly 50/50 split between optimal and suboptimal trajectories, so it saw greater optimal performance earlier on. Because participants directly teleoperate the robot, active tasks are noisier. This means that many trajectories are suboptimal, with some failing to reach the goal at all. Pooled data combines both interaction tasks, and therefore trains faster than Active alone but slower than Passive alone.

\subsection{Ablation}
Finetune checkpoint ablations show that early injection yields the best performance across all conditions. A threshold of 0.2 shows some benefit, but beyond that checkpoint, the neural signal does not improve learning. 

We ran an ablation study to test whether performance gains stem from the neural signal, instead of artifacts from the framework. Conditions with considerable gains (Q-Augmentation) show worsened performance when more noise is added to the performance decoder. For conditions without considerable gains (Reward Augmentation), adding noise has little to no effect (Fig. \ref{fig:noise}). A pattern evolves, supporting the idea that performance improvements are driven directly by the neural signal instead of by artifacts in the framework. 

\section{Conclusion and Future Work}
This preliminary work suggests that fNIRS signals can be used to update a robot learning algorithm and improve training efficiency. We introduce \textit{NEURO-LOOP}, a framework that evaluates various augmentation methods, compares decoder granularity, and tests robustness to noise and finetuning checkpoints. Our results demonstrate that this pipeline effectively uses offline data. We believe that \textit{NEURO-LOOP} may provide a useful starting point for future work extending this approach to real-time human-robot interaction settings.

There are several limitations to this research that we believe should be tackled in future work. First, agent performance labels are a function of the agent's optimality, and they are not guaranteed to reflect the participant's true judgment. We assume that the participant's judgment of the agent aligns with that of a near-optimal policy, however we believe that exploring labeling schemes to capture user preference more directly will benefit these fNIRS-based systems. Second, our decoders avoid the credit assignment problem by using 4-second neural feature windows. Future work should explore using standardized methods to account for temporal delay between the fNIRS hemodynamic response and the interaction with the agent at that time. Finally, further work should compare this method against other human-in-the-loop (HITL) approaches. We also believe that extending these methods to an online setting and to a physical robot are natural next steps.

\section*{Acknowledgments}
We thank our colleagues at Tufts University, the MuLIP Lab, and the HCI lab for their feedback and support at each step of this project.

\section*{Code \& Datasets}
\noindent
\textbf{Code: } \url{https://github.com/your-profile/OfflineNeuroloop/tree/iros2026}
\\
\textbf{Dataset: } \url{https://github.com/your-profile/fNIRS2RL}





\vspace{12pt}
\end{document}